%% file: paper.tex
\documentclass[sigconf,nonacm]{acmart}
\setkeys{acmart.cls}{balance=false}
\pdfmapfile{+libertine.map}
\microtypesetup{expansion=false}

\setcopyright{none}
\usepackage[T1]{fontenc}
\usepackage[utf8]{inputenc}
\usepackage{microtype}
\usepackage{graphicx}
\usepackage{booktabs}
\usepackage{amsmath}
\usepackage{subcaption}
\usepackage{array}
\usepackage{tikz}
\usetikzlibrary{arrows.meta,positioning,fit,backgrounds,calc,shapes.geometric}
\usepackage{siunitx}
\usepackage[capitalize]{cleveref}
\usepackage{algorithm}
\usepackage{algpseudocode}
\usepackage{needspace}

\title{CLIPPER Beyond Shortlisting: Auditable Decision Support for Changing Municipal Micromobility Policies}
\author{Julian Teusch}
\email{julian.teusch@tu-clausthal.de}
\affiliation{%
  \department{Institute of Computer Science}
  \institution{Clausthal University of Technology}
  \city{Clausthal-Zellerfeld}
  \country{Germany}
}
\author{Jörg Philipp Müller}
\email{joerg.mueller@tu-clausthal.de}
\affiliation{%
  \department{Institute of Computer Science}
  \institution{Clausthal University of Technology}
  \city{Clausthal-Zellerfeld}
  \country{Germany}
}
\author{Monika Sester}
\email{monika.sester@ikg.uni-hannover.de}
\affiliation{%
  \department{Institute of Cartography and Geoinformatics}
  \institution{Leibniz University Hannover}
  \city{Hannover}
  \country{Germany}
}
\renewcommand{\shortauthors}{Teusch et al.}

\begin{document}
\raggedbottom

\begin{abstract}
In municipal planning workshops, planners and other stakeholders compare shared-micromobility
parking policies by varying no-parking zones, retained sites, spacing, or area allocations. Each edit
changes feasible sites and how much demand they cover, so the
alternative must be reoptimized on the same spatial data. Full-set greedy, the transparent reference
for this task, takes tens of seconds per alternative at city scale. We
present Constraint-exact Low-latency Iterative Planning with Pooled Evaluation and Replay
(\textbf{CLIPPER}), an optimizer with audit functions developed for requirements elicited with
the City of Braunschweig. In each
greedy round, it forms a deterministic candidate pool of bounded size, computes how much
still-uncovered demand each candidate would add, and rejects candidates that violate an active
constraint. An optional offline audit scans every remaining feasible candidate and records what the
restricted pool omitted. We evaluate these functions on complete eleven-state edit chains (\(E_0,\ldots,E_{10}\))
in Braunschweig, Munich, and Berlin. With \(K=1024\) candidates per group, the fixed-width mode
CLIPPER-F has mean coverage gaps to full-set greedy under the same policy of \(0.245\), \(0.003\),
and \(0.001\) percentage points in Braunschweig, Munich, and Berlin, respectively, while mean
rollout time falls by factors of \(13.6\text{--}28.9\); no audited run terminates while a candidate
outside the pool could still increase coverage. Plans computed from two checksummed versions of
Braunschweig's official no-parking-zone data differ in 30 of about 540 selected sites although coverage moves by only about
\(0.1\) percentage points. These changes still require municipal assessment and implementation. The
findings inform a proposed municipal process that versions policy inputs, reports site changes beside
coverage, and scans the full candidate set before a final decision.
\end{abstract}

\ccsdesc[500]{Information systems~Decision support systems}
\ccsdesc[300]{Theory of computation~Submodular optimization and polymatroids}
\ccsdesc[300]{Applied computing~Transportation}

\keywords{spatial decision support, facility location, submodular coverage, auditability}

\maketitle
\begingroup
\renewcommand{\thefootnote}{}
\footnotetext[0]{Author preprint. Accepted for publication in the proceedings of the 2nd ACM SIGSPATIAL International Workshop on Spatial Intelligence for Smart and Connected Communities (SpatialConnect 2026).}
\endgroup

\input{sections/introduction.tex}

\input{sections/related_work.tex}

\input{sections/problem_formulation.tex}

\input{sections/clipper_audit_ready_shortlisted_execution.tex}

\input{sections/empirical_evaluation.tex}

\input{sections/implementation_pathway.tex}

\Needspace{8\baselineskip}
\input{sections/discussion.tex}

\input{sections/conclusion.tex}

\bibliographystyle{ACM-Reference-Format}
\bibliography{references}

\end{document}

%% file: sections/introduction.tex
\section{Introduction}

Shared micromobility can support first/last-mile access, but without deliberate parking and station
planning it also creates curb-space and compliance problems
\cite{Shaheen_2022,Yin_2024,Klein_2023}. A municipality designating shared-scooter parking must
choose a limited number of zones so that as many trips as possible begin or end within walking
distance of one. We call each possible zone location a \emph{candidate}. Policy rules narrow the
choice. \emph{Exclusions} close areas to new parking, \emph{locks} retain existing sites, and
\emph{spacing} keeps new zones a minimum walking distance apart. \emph{Caps} distribute the facility
budget across areas by limiting how much any one area may receive.

Workshop-based comparison turns these rules into alternatives: municipal planners and other
stakeholders vary them, inspect a feasible plan after each edit, and compare coverage and site
changes. Braunschweig's municipal quality agreement permits riding and parking restrictions to be
revised at any time, including temporary no-return areas for major events
\cite{StadtBraunschweig_2021}. Our collaboration with the City established requirements to revise
restrictions, retain mandatory sites, enforce spacing and area allocations, and compare feasible
alternatives over common data. We encode these requirements in the scenario model. Each edit changes
admissible candidates, so the previous plan can become infeasible or cover less demand. The task is
therefore not one-off facility placement, but repeated comparison of policy alternatives on a
common spatial basis.

Comparison must also reach below aggregate numbers. In our documented policy-input case, two
official versions of Braunschweig's no-parking-zone data, each identified by a checksum, produce
plans that differ in 30 of about \num{540} selected sites although aggregate coverage moves by only
about \(0.1\) percentage points (\cref{tab:policy-snapshots}). A municipality would still have to
assess, implement, and communicate every such change, so decision support for changing policies
must version its inputs and report which selected sites change, not only how much demand is
covered.

\paragraph{From one optimization run to repeated comparison.}
Planning-support research treats rapid ``what-if'' exploration as an aid to deliberation rather than
a replacement for planning judgment \cite{Klosterman_1997,Pelzer_2014,Pettit_2018}. A repeated
edit--solve--compare loop creates four computational requirements. Every returned plan must satisfy
the active spatial and policy constraints. Solving each alternative must be fast enough to make a
sequence of comparisons practical. If the optimizer considers only part of the candidate set, the
effect of that restriction must be measurable. Finally, stored scenarios and deterministic execution
must allow a result to be reproduced and compared later. We evaluate these properties on
complete, controlled scenario chains and on documented municipal policy inputs.

\paragraph{Technical gap.}
Maximal covering and monotone submodular optimization provide transparent siting models
\cite{Church_ReVelle_1974,Nemhauser_1978}. \emph{Full-set greedy} repeatedly adds the feasible
candidate with the largest current coverage gain. We use it as the reference because it applies the
same transparent selection rule to the complete candidate set under the same constraints and caps,
thereby isolating the effect of restricting the pool on coverage. In our
benchmarks, this control averages \(23\text{--}53\) seconds per state, so all eleven states require
\(4.2\text{--}9.7\) minutes of solver time before inspection or discussion. Lazy,
stochastic, and thresholded greedy methods reduce the number of gain calculations under their
respective models \cite{Badanidiyuru_Vondrak_2014,Mirzasoleiman_2015,Krause_Golovin_2014}; dynamic
and prediction-augmented methods maintain solutions across update sequences
\cite{Agarwal_Balkanski_2024}. CLIPPER solves every stored state from the beginning. Its selector
enforces locks, exclusions, area caps, conflict classes that prevent duplicate use of the same curb
space, and a minimum network distance between sites. A replaceable rule limits the candidates
considered in each round. Within that pool, the
selector computes exact current coverage gains and checks every active constraint before adding a
site.

\paragraph{Contribution.}
CLIPPER narrows \emph{what the optimizer looks at} without relaxing \emph{what it enforces}. We
contribute Constraint-exact Low-latency Iterative Planning with Pooled Evaluation and Replay
(CLIPPER) as an optimizer with audit functions for municipal decision support. Its contributions are:
\begin{itemize}
  \item \textbf{Exact enforcement of the recorded policy.} The selector starts with mandatory locks
    and adds only candidates that satisfy the active exclusions, spacing rule, conflict classes,
    area caps, and global budget. A fixed ordering rule resolves ties, so the same recorded inputs
    produce the same selections.
  \item \textbf{Two explicit ways to form the candidate pool.} In the fixed-width mode
    CLIPPER-F(\(K\)), every active proposal group offers up to \(K\) candidates under balanced area
    caps. The adaptive-budget mode CLIPPER-A(\(B_{\mathrm{prop}}\)) distributes one total candidate
    budget among the groups and uses
    a stated policy that gives more capacity to areas with more active candidates. Each mode is
    compared with full-set greedy under that mode's policy.
  \item \textbf{Separate checks during planning and after a run.} A lightweight score can request a
    larger pool or a later audit. The exact offline audit instead scans every remaining feasible
    candidate and records how much additional coverage the restricted pool omitted in each round.
  \item \textbf{Municipally grounded requirements and policy evidence.} Two versions of the city's
    official no-parking zones show how revised exclusion geometry changes both feasibility and the
    selected sites; each version is identified by a checksum
    \cite{StadtBraunschweig_Parkverbotszonen_2026}. These results
    support a proposed municipal process from registering policy data and calibrating the candidate
    rule to comparing alternatives, making a final decision, and archiving its audit
    (\cref{sec:implementation}).
\end{itemize}

\paragraph{Extension beyond the four-page report.}
An accepted four-page SIGSPATIAL paper introduced CLIPPER and the primary comparisons over the
eleven-state chains \cite{Teusch_2026_CLIPPERShort}. Beyond that short paper, this article adds four
elements: audits that scan all remaining candidates at three values of \(K\), a comparison rule that
updates scores only for candidates affected by newly covered demand, an evaluation using two official
versions of municipal policy geometry, and the implementation process in \cref{sec:implementation}.
We test the optimization procedure on all eleven states (\(E_0,\ldots,E_{10}\)) of controlled
three-city chains and examine revised municipal geometry in the Braunschweig case.

We evaluate these technical functions through three questions, then set out their proposed use in
municipal planning (\cref{sec:implementation}):
\begin{itemize}
  \item[\textbf{Q1}] How much time does the restricted pool save under the same policy, how large
    is the remaining coverage gap, and does the returned plan satisfy every encoded constraint?
  \item[\textbf{Q2}] How often does a pool ordered once per state omit a larger feasible gain,
    and how do score updates after each selection change coverage
    and runtime?
  \item[\textbf{Q3}] How much can two official versions of exclusion geometry change the selected
    sites when their aggregate coverage is similar?
\end{itemize}
For Q1, CLIPPER-F(\(1024\)) has a mean coverage gap over the eleven states of \(0.245\) percentage
points in Braunschweig, \(0.003\) in Munich, and \(0.001\) in Berlin, and cuts mean rollout time
by factors of \(13.6\text{--}28.9\). Under its coverage-prioritized policy, CLIPPER-A(\(8192\)) differs from the
corresponding full-set greedy run by \(1.82\) percentage points in Braunschweig, \(0.12\) in Munich,
and \(0.27\) in Berlin, while using \(9\text{--}15\%\) of that run's rollout time
(\cref{tab:method-comparison-main,sec:evaluation}). Q2 and Q3 are answered by the round-by-round
audit, score-update comparison, and municipal policy-input case in
\cref{tab:proposal-order-audit,tab:policy-snapshots}. We additionally verify deterministic replay as
a system property in \cref{sec:cert-empirical}.

%% file: sections/related_work.tex
\section{Related work}

\paragraph{Planning support and repeated comparison.}
Planning support systems organize ``what-if'' analysis, communication about scenarios, and
stakeholder deliberation \cite{Klosterman_1997,Geertman_2015_PSSSmartCities}. Research on their
adoption emphasizes usability, trust, and transparent trade-offs rather than autonomous replacement
of planning judgment \cite{Pelzer_2014,Russo_2018,Pettit_2018}. CLIPPER supplies the computational
part of such a workflow: it solves each recorded policy state independently, retains the information
needed to compare that result later, and measures the effect of restricting the candidate pool.

\paragraph{Reducing work in greedy submodular optimization.}
Maximal covering and monotone submodular maximization provide the objective foundation
\cite{Church_ReVelle_1974,Nemhauser_1978}. Standard greedy repeatedly selects the feasible candidate
with the largest current gain. Lazy evaluation keeps the full candidate set but uses valid upper
bounds to avoid recalculating every gain. Stochastic greedy samples candidates in each round.
Thresholded methods skip candidates whose gains fall below a decreasing threshold
\cite{Krause_Golovin_2014,Mirzasoleiman_2015,Badanidiyuru_Vondrak_2014}. These approaches reduce the
number of gain calculations under their stated feasibility models. CLIPPER limits the set passed to
the selector, calculates exact current gains within that set, and checks the combined
scenario constraints before every insertion. The evaluation includes full-set, stochastic,
thresholded, and local-search controls. Its primary comparisons use full-set greedy under the same
policy as the corresponding CLIPPER mode.

\paragraph{Optimization under changing inputs.}
Dynamic submodular algorithms maintain a solution as elements or objective information change;
learning-augmented variants use forecasts to reduce the work needed for an update
\cite{Agarwal_Balkanski_2024}. Agarwal and Balkanski study updates under a cardinality constraint,
whereas the planning states here combine locks, caps for individual areas, conflict classes,
exclusions, and network-distance spacing. CLIPPER discards the previous solution after a policy edit
and solves the recorded state from the beginning; \cref{sec:evaluation} additionally compares
candidate orders that do and do not update scores during a solve.

\paragraph{Facility location for micromobility.}
Station-location studies for bicycle sharing use geographic information systems (GIS) and demand
data
\cite{GarciaPalomares_2012}, including crowd-planned station redeployment in SIGSPATIAL
\cite{Zhang_2016}. More recent work considers geofenced parking identification, station planning,
and time-varying curb allocation \cite{Zhao_Ong_2021,Cai_2023_Geofencing,Schwerdfeger_2025}.
Reinforcement learning has also been applied to strategic geofenced facility planning
\cite{Teusch_2025}. CLIPPER focuses on deterministic recomputation of weighted coverage plans under
an explicit policy state.

\paragraph{Learning for combinatorial optimization.}
Learned candidate rules can accelerate combinatorial optimization, while hard constraints and
distribution shift remain central deployment concerns \cite{Bengio_2021,Kotary_2021}. CLIPPER
accepts a replaceable source for its candidate pool. Throughout the experiments, the deterministic
score is the demand that each candidate would cover on its own.

%% file: sections/problem_formulation.tex
\section{Problem formulation}
\label{sec:problem}

This section defines one recorded policy state, its feasible plans, and the coverage objective;
\cref{sec:method} then describes how CLIPPER computes a plan.

\paragraph{Candidates and demand.}
Let \(V\) be the set of unique candidate facility locations (parking points or areas). Demand is
represented by weighted points \(d\in\mathcal D\) with weights \(w_d\ge 0\). Each candidate
\(e\in V\) covers a subset \(C_e\subseteq\mathcal D\), such as demand within a walk-access radius.
For a selected plan \(S\subseteq V\), the weighted union coverage is
\begin{equation}
f(S) = \sum_{d\in\mathcal{D}} w_d \cdot \mathbf{1}\left[d \in \bigcup_{e\in S} C_e\right],
\label{eq:coverage-objective}
\end{equation}
where \(\mathbf 1[\cdot]\) is the indicator function. The selector, the audit, and the reported
coverage values all derive from \(f\).
Monotonicity and submodularity follow directly from the coverage form: the marginal contribution
of a candidate is the nonnegative weight of still-uncovered demand in \(C_e\), which can only shrink
as the incumbent set grows.

\paragraph{Proposal groups and accounting groups.}
Pool construction and cap accounting serve different roles and may use different groupings. Each candidate belongs to exactly one
\emph{accounting group}, whose cap counts that candidate. The index set
\(\mathcal G_{\mathrm{acc}}\) and assignment \(a:V\rightarrow\mathcal G_{\mathrm{acc}}\) define the
disjoint accounting sets \(V_h^{\mathrm{acc}}:=\{e\in V:a(e)=h\}\). The \emph{proposal groups} used
to form the candidate pool
have index set \(\mathcal G_{\mathrm{prop}}\) and candidate sets
\(V_g\subseteq V\), \(g\in\mathcal G_{\mathrm{prop}}\). Proposal groups may overlap, so the same
physical location may enter the pool through more than one group; its accounting assignment remains
unique.

\paragraph{Scenario configuration: spatial and policy constraints.}
The planning scenario \(\Omega\) records exclusions, locks, spacing, budgets, and accounting-group
caps. It specifies the active proposal groups
\(A(\Omega)\subseteq\mathcal G_{\mathrm{prop}}\), forbidden candidates
\(Z(\Omega)\subseteq V\), and locked facilities \(L(\Omega)\subseteq V\). The global facility cap is
\(B(\Omega)\), and \(U_h(\Omega)\) is the cap for accounting group
\(h\in\mathcal G_{\mathrm{acc}}\). Duplicate/conflict classes
\(\mathcal H(\Omega)\subseteq 2^V\) group candidates
that describe the same physical curb space; at most one candidate in each class may be chosen. Let
\(\mathrm{dist}(e_i,e_j)\) denote the configured network distance between two candidates. The
minimum permitted distance \(d_{\min}(\Omega)\ge 0\) defines the spacing rule.
An edit changes one or more of these spatial and policy components, and the resulting scenario
\(\Omega\) is solved and audited as a new state. In the benchmark, the conflict classes are
disjoint; the distance rule handles arbitrary pairwise spacing conflicts separately.
The candidates available for \emph{new} placements are
\[
  V^{\mathrm{new}}(\Omega):=\left(\bigcup_{g\in A(\Omega)}V_g\right)\setminus Z(\Omega).
\]
Given \(\Omega\), let \(\mathcal I(\Omega)\) denote the feasible family
\begin{equation}
\mathcal I(\Omega):=\left\{S\subseteq V:
\begin{aligned}
& L(\Omega)\subseteq S, \\
& S\setminus L(\Omega) \subseteq V^{\mathrm{new}}(\Omega), \\
& |S|\le B(\Omega), \\
& |S\cap V_h^{\mathrm{acc}}|\le U_h(\Omega)\ \ \forall h\in\mathcal G_{\mathrm{acc}}, \\
& |S\cap H|\le 1\ \ \forall H\in\mathcal H(\Omega), \\
& \mathrm{dist}(e_i,e_j)\ge d_{\min}(\Omega)\ \ \forall e_i\neq e_j\in S.
\end{aligned}
\right\},
\label{eq:problem}
\end{equation}
and seek \(\max_{S\in\mathcal I(\Omega)} f(S)\).
Under our scenario semantics, \emph{locks override exclusions}. Locked facilities \(L(\Omega)\)
must appear in every feasible solution even if they fall inside exclusion zones, since they
represent existing infrastructure commitments. Exclusions apply only to \emph{new} placements
(\(S\setminus L(\Omega)\)). We assume that the locks themselves form a feasible plan,
\(L(\Omega)\in\mathcal I(\Omega)\). We fix these mandatory locks before analyzing the remaining
choices. The residual global budget is \(B^L(\Omega):=B(\Omega)-|L(\Omega)|\), and the residual
candidate set and feasible family are
\begin{align*}
  V^{L}(\Omega)&:=V^{\mathrm{new}}(\Omega)\setminus L(\Omega),\\
  \mathcal I^{L}(\Omega)&:=\{X\subseteq V^{L}(\Omega):
    L(\Omega)\cup X\in\mathcal I(\Omega)\}.
\end{align*}
The selector operates on \(\mathcal I^L(\Omega)\) and returns \(L(\Omega)\cup X\) for the original
problem. In the evaluation, every recorded edit produces a new instance that is solved from the
beginning.

\paragraph{Temporal aggregation.}
Trip records are aggregated into weighted demand points, and the optimizer solves the resulting
spatial objective in \eqref{eq:coverage-objective}. Temporal utilization, rebalancing, and
peak-period service can be analyzed after spatial selection
\cite{Shaheen_2022,Teusch_2023,Schleibaum_2025}; \cref{sec:evaluation} defines the benchmark
construction scope used for the tables in this paper.

%% file: sections/clipper_audit_ready_shortlisted_execution.tex
\section{CLIPPER: optimization over a restricted candidate pool}
\label{sec:method}

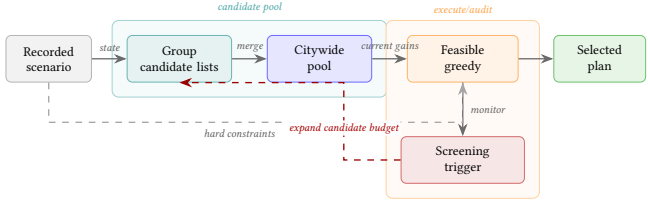
\begin{figure}[t]
  \centering
  \resizebox{\columnwidth}{!}{%
  \begin{tikzpicture}[
    >=Stealth,
    node distance=4.4mm and 5.5mm,
    every node/.style={font=\scriptsize},
    box/.style={draw, rounded corners=2.4pt, minimum height=7.4mm,
                text width=15.5mm, align=center, fill=#1!10, draw=#1!58,
                line width=0.45pt, inner sep=1.5pt},
    box/.default=blue,
    group/.style={draw=#1!36, fill=#1!4, rounded corners=3pt, inner sep=2.3mm},
    lbl/.style={font=\tiny\itshape, text=black!62},
    lblbg/.style={font=\tiny\itshape, text=black!78, fill=white, fill opacity=0.94,
                  text opacity=1, rounded corners=1.4pt, inner sep=1.1pt},
    arr/.style={->, semithick, draw=black!58},
  ]
    \node[box=gray, text width=12.5mm] (omega) {Recorded\\scenario};
    \node[box=teal, right=of omega] (sl) {Group\\candidate lists};
    \node[box=blue, right=of sl] (pool) {Citywide\\pool};
    \node[box=orange, right=of pool, text width=16.5mm] (exec) {Feasible\\greedy};
    \node[box=green!60!black, right=of exec, text width=13.5mm] (out) {Selected\\plan};
    \node[box=red!70!black, below=8.5mm of exec, text width=18.5mm] (cert) {Screening\\trigger};

    \draw[arr] (omega) -- node[above, lbl] {state} (sl);
    \draw[arr] (sl) -- node[above, lbl] {merge} (pool);

    \draw[arr] (pool) -- node[above, lbl] {current gains} (exec);
    \draw[arr] (exec) -- (out);

    \draw[arr] (exec) -- node[right, lbl] {monitor} (cert);
    \draw[arr, dashed, draw=gray!70]
      (omega.south) -- ++(0,-6.3mm) -| node[pos=0.23, below, lbl] {hard constraints} (exec.south);

    \draw[arr, dashed, draw=red!60!black]
      (cert.west) -- ++(-9mm,0) |- node[pos=0.26, below, lblbg, text=red!65!black] {expand candidate budget} (sl.south);

    \begin{scope}[on background layer]
      \node[group=teal, fit=(sl)(pool), label={[lbl, teal!70]above:candidate pool}] {};
      \node[group=orange, fit=(exec)(cert), label={[lbl, orange!72]above:execute/audit}] {};
    \end{scope}
  \end{tikzpicture}%
  }
  \caption{CLIPPER pipeline. Scenario constraints govern both the candidate pool and greedy
  selection. The screening rule (defined later in this section) can request a larger pool or an
  offline audit. That audit scans all remaining candidates and calculates the coverage gain omitted
  from the restricted pool.}
  \Description{CLIPPER pipeline diagram: scenario constraints feed per-group candidate rules that form a pooled candidate set, a restricted-pool greedy selector selects feasible actions under constraint checks, and an optional screening rule can request candidate-budget expansion.}
  \label{fig:clipper-pipeline}
\end{figure}

A planning session poses \eqref{eq:problem} again after every edit, so the cost that matters is the
time to solve one recorded state. CLIPPER can use different rules to decide which candidates the
greedy selector evaluates. Its core loop separates three tasks
(\cref{fig:clipper-pipeline,alg:clipper}). First, the candidate rule forms a pool from the active
proposal groups. Second, the selector computes each pooled candidate's exact \emph{current marginal
gain}
\(\Delta(e\mid S):=f(S\cup\{e\})-f(S)\): the weight of still-uncovered demand that candidate
\(e\) would add to the plan \(S\) built so far.
It adds the candidate with the largest gain among those that satisfy every active constraint. Third,
an optional offline audit compares this restricted pool with all remaining feasible candidates. The
experiments use deterministic pools ordered by singleton coverage, which is the demand a candidate
would cover if selected on its own.

Two operating modes share the same selector. The fixed-width mode CLIPPER-F(\(K\)) offers at most
\(K\) candidates from each active proposal group and uses balanced accounting caps. The
adaptive-budget mode CLIPPER-A(\(B_{\mathrm{prop}}\)) splits a total pool budget of
\(B_{\mathrm{prop}}\) candidates among the proposal groups and uses the coverage-prioritized
accounting caps defined below. This pool budget is separate from the facility budget \(B(\Omega)\)
that bounds the final plan (\cref{eq:problem}). \Cref{alg:clipper} covers both modes through a
configurable rule for allocating candidate slots; CLIPPER-F sets \(K_{g,t}=K\) for every active
group.

For each proposal group \(g\), let \(\pi_g\) be its stable candidate order. In round \(t\), the pool
rule assigns \(K_{g,t}\) slots to group \(g\), which offers
a list \(P_{g,t}\); the selector evaluates the deduplicated citywide pool
\(P_t:=\bigcup_{g\in A(\Omega)}P_{g,t}\). We write \(\mathcal I\) for
\(\mathcal I(\Omega)\) where unambiguous. The set \(S_t\) contains the residual unlocked selections
after \(t\) rounds, and \(S_t^+:=L(\Omega)\cup S_t\) is the corresponding full solution including
locks. Let \(T\) be the number of completed selection rounds.

\begin{algorithm}[t]
  \caption{CLIPPER with a configurable rule for allocating candidate slots.}
  \label{alg:clipper}
  \footnotesize
  \begin{algorithmic}[1]
    \Require Encoded scenario $\Omega$; active proposal groups $A(\Omega)$; candidates $\{V_g\}$; stable group orders $\{\pi_g\}$; rule \(\mathsf{Budget}\) for allocating candidate slots; optional screening rule \(\mathsf{Trigger}\)
    \Ensure Feasible selected plan including mandatory locks
    \State $S \gets \emptyset$ \Comment{residual selections; locks fixed separately}
    \State $B^L(\Omega) \gets B(\Omega)-|L(\Omega)|$ \Comment{residual global budget}
    \For{$t=1,2,\dots,B^L(\Omega)$}
      \State $\{K_{g,t}\}_{g\in A(\Omega)} \gets \mathsf{Budget}(\Omega,S,t)$
      \ForAll{$g\in A(\Omega)$}
        \State $P_{g,t}\gets$ first $K_{g,t}$ currently valid, unselected candidates in $\pi_g$, continuing past skipped candidates
      \EndFor
      \State $P_t \gets \bigcup_{g\in A(\Omega)} P_{g,t}$
      \State $S^+ \gets L(\Omega)\cup S$
      \State $F_t \gets \{e\in P_t:\; S^+\cup\{e\}\in\mathcal I(\Omega)\}$
      \If{$F_t=\emptyset$}
        \State Enlarge the pool and restart round \(t\), or \textbf{break}, marking the stop unverified unless a full residual scan classifies it
      \EndIf
      \State $e_t \gets \arg\max_{e\in F_t}\bigl[f(S^+\cup\{e\})-f(S^+)\bigr]$ \Comment{first in pool order breaks exact ties}
      \State Evaluate \(\mathsf{Trigger}(\Omega,S^+,P_t)\) \Comment{optional screening rule}
      \If{$\mathsf{Trigger}$ requests a larger pool}
        \State Enlarge the pool and restart round \(t\)
      \EndIf
      \If{$f(S^+\cup\{e_t\})-f(S^+)\le 0$}
        \State Enlarge the pool and restart round \(t\), or \textbf{break}, marking the stop unverified unless a full residual scan classifies it
      \EndIf
      \State $S \gets S\cup\{e_t\}$
    \EndFor
    \State \Return $L(\Omega)\cup S$
\end{algorithmic}
\end{algorithm}
A stop is \emph{truncated} if the pool has no positive feasible action but a full residual scan
finds one outside it; otherwise the scan confirms termination. Without that scan, the stop is
\emph{unverified}. The experiments report audit-confirmed truncations separately from coverage.

For deterministic replay (re-running a recorded state to reproduce the same plan), the recorded
context includes the scenario, demand and candidate data, proposal-group order, candidate ranks,
allocation and cap policies, executor settings, and software version. When gains are exactly equal, the selector takes the first candidate in the deduplicated
pool's deterministic order. This order follows the recorded proposal-group order and the stable rank
within each group.

\paragraph{Restricted candidate pools.}
At each step \(t\), each active proposal group \(g \in A(\Omega)\) produces a candidate list
\(P_{g,t}\subseteq V_g\cap V^L(\Omega)\) with \(|P_{g,t}| \le K_{g,t}\) using a ranking source
under a common interface. Here \(K_{g,t}\) limits one group's list; it is not the size of the
citywide pool after the lists are combined. For the weighted coverage benchmark, CLIPPER-F and
CLIPPER-A rank candidates by their deterministic singleton coverage
\(\operatorname{pot}_{\Omega}(e):=f(\{e\})\), restricted to candidates currently feasible under the
scenario filters. Because
\[
  \Delta(e\mid S)=\sum_{d\in C_e\setminus \cup_{j\in S} C_j} w_d
  \le \sum_{d\in C_e} w_d
  = f(\{e\})
  = \operatorname{pot}_{\Omega}(e),
\]
the same score gives a deterministic order and is an upper bound on the candidate's current gain.

To build a group list, the implementation scans each proposal group's stable rank order. It skips
already selected candidates \(S_{t-1}\) and candidates that fail the current cap, conflict, or
spacing checks against \(S_{t-1}^{+}\). It continues through the same order until reaching the
requested width or exhausting the group. Thus
``top-\(K\)'' means the first \(K\) valid candidates in the deterministic order: when an invalid
candidate is skipped, scanning continues so that its place can be filled.
Other ranking sources use the same interface; changing the ranking source leaves the selector,
feasibility, and audit definitions unchanged.

\paragraph{Greedy selection over the restricted pool.}
With residual selections \(S_{t-1}\) and the full plan including locks
\(S_{t-1}^+:=L(\Omega)\cup S_{t-1}\), the algorithm forms the pool
\(P_t = \bigcup_{g\in A(\Omega)} P_{g,t}\) and selects the best feasible action by evaluating the
exact marginal gain in the citywide coverage objective:
\begin{equation}
e_t \in \arg\max_{\substack{e \in P_t\setminus S_{t-1} \\ S_{t-1}^+\cup\{e\}\in\mathcal I}}
\Delta(e\mid S_{t-1}^+).
\end{equation}
Feasibility is enforced by checks derived from $\Omega$ (budgets, exclusions, spacing,
duplicate/conflict classes). Locked facilities remain included regardless of exclusion zones. If a
candidate rule suggests an invalid candidate, the selector rejects it.

\paragraph{Residual view with mandatory locks.}
Because locks are compulsory elements, the formal analysis is carried out on the residual instance
obtained after fixing them. It uses the residual family \(\mathcal I^{L}(\Omega)\) from
\cref{sec:problem} and the residual objective
\(f^{L}_{\Omega}(X):=f(L(\Omega)\cup X)-f(L(\Omega))\) on \(V^{L}(\Omega)\).
Fixing mandatory elements preserves monotone submodularity. The selector is therefore greedy on the
residual instance, initialized at the empty set and adding only unlocked facilities; its reported
solution in the original problem is \(S_T^+:=L(\Omega)\cup S_T\).

\paragraph{Offline scan of all remaining candidates.}
Let \(S_{t-1}^{+}:=L(\Omega)\cup S_{t-1}\) denote the full plan after \(t-1\) residual selections and let
\(R_{t-1}(\Omega):=V^L(\Omega)\setminus S_{t-1}\) be the unselected residual candidates. At each
round, the audit compares the largest feasible gain among all remaining candidates with the largest
gain visible in the restricted pool. Define
\[
  F_t^{\mathrm{all}}:=\{e\in R_{t-1}(\Omega):
  S_{t-1}^{+}\cup\{e\}\in\mathcal I(\Omega)\},
  \qquad F_t^{\mathrm{pool}}:=F_t^{\mathrm{all}}\cap P_t.
\]
The two gains are
\begin{equation}
\begin{aligned}
  m_t^*&:=\max\bigl(\{\Delta(e\mid S_{t-1}^{+}):e\in F_t^{\mathrm{all}}\}\cup\{0\}\bigr),\\
  m_t&:=\max\bigl(\{\Delta(e\mid S_{t-1}^{+}):e\in F_t^{\mathrm{pool}}\}\cup\{0\}\bigr).
\end{aligned}
\end{equation}
Including zero defines both quantities even when one of the sets is empty. The difference
\(\delta_t:=m_t^*-m_t\) is the coverage gain omitted from the pool in round \(t\). If
\(\delta_t=0\), the pool omits no larger feasible gain. If also \(m_t^*>0\), it contains a candidate
attaining that gain; at zero gain, it may be empty. At a terminal check, \(m_t=0<m_t^*\) means that the pool
would stop even though a candidate outside it could still improve coverage. Computing \(m_t^*\)
requires exact gains for the full remaining set, so this audit is used after a run or at a selected
decision checkpoint.

\paragraph{Singleton-coverage screening score.}
During planning, the screening rule (\(\mathsf{Trigger}\) in \cref{alg:clipper}) can use the less
expensive singleton upper bound
\[
u_t:=\max\bigl(\{\operatorname{pot}_{\Omega}(e):e\in F_t^{\mathrm{all}}\}\cup\{0\}\bigr),
\]
which satisfies \(u_t\ge m_t^*\). Therefore
\(\delta_t\le\hat\delta_t^{\mathrm{ub}}:=u_t-m_t\). This screening score is a conservative upper
bound on the gain omitted in the current round and can trigger a larger pool or an offline audit. It does not
bound the coverage difference between the final plans. To certify a stop before exhausting the residual
budget, verify \(u_{T+1}\le 0\) or scan all remaining candidates. For offline
reporting, the audit can sum \(\delta_t\) over all audited rounds.

\paragraph{Operating policies: balanced caps and coverage-prioritized caps.}
The \emph{balanced-cap policy} gives every accounting group the benchmark's equal cap and is used by
CLIPPER-F and its corresponding full-set greedy run. The planner-specified
\emph{coverage-prioritized policy} keeps
the global budget unchanged but assigns more of it to groups with more active candidates. After
subtracting locks, largest-remainder rounding divides the remaining budget in proportion to active
candidate counts. With multiplier \(\lambda=1.5\), CLIPPER-A records the cap
\(U_h^A=\ell_h+\lceil \lambda b_h\rceil\), where \(\ell_h\) is the number of locked facilities and
\(b_h\) is that group's rounded share. These caps are stored in the scenario before selection and
enforced as hard constraints. CLIPPER-A and its corresponding full-set greedy run both use them.

\paragraph{Budgeted adaptive candidate policy.}
CLIPPER-A keeps the deterministic singleton-coverage ranking rule and exact greedy selection, and
replaces the fixed number per group with \(B_{\mathrm{prop}}\) total slots per round. Every selected
candidate \(e\) is charged to its accounting assignment \(a(e)\). For allocating slots in the
evaluated benchmark, each proposal group \(g\) is aligned with one accounting group \(h(g)\), with
\(V_g\subseteq V_{h(g)}^{\mathrm{acc}}\). A group receives no more slots once that cap is full. Among groups whose
caps are not full, the allocation depends on the number of feasible candidates and the highest
singleton score. It first reserves the configured minimum number of slots for each such group, then
uses largest-remainder rounding and resolves ties by recorded group order. We evaluate
\(B_{\mathrm{prop}}\in\{4096,8192\}\); the main CLIPPER-F comparison uses \(K=1024\)
(\cref{tab:method-comparison-main}). Both modes share the objective, data, selector, and constraint
checks but use the policies and slot allocations stated above.

\paragraph{Complexity and cost comparison.}
Let \(c:=\max_{e\in V^L(\Omega)}|C_e|\) be the largest demand support of one candidate. The number
of completed rounds satisfies \(T\le B^L(\Omega)\). If \(N_t\) candidates reach the selector in round
\(t\), CLIPPER performs \(O(c\sum_{t=1}^T N_t)\) work for gain calculations. For CLIPPER-F this is
\(O(TG_{\mathrm{eff}}Kc)\), where \(G_{\mathrm{eff}}\) is the number of active groups; for
CLIPPER-A it is \(O(TB_{\mathrm{prop}}c)\). Full-set greedy uses
\(O(T|V^L(\Omega)|c)\). If the pool contains every remaining feasible candidate in every round,
\(\delta_t=0\) and CLIPPER returns the same greedy sequence under the stated tie rule.

%% file: sections/empirical_evaluation.tex
\section{Empirical evaluation}
\label{sec:evaluation}

\begin{figure*}[t]
  \centering
  \captionsetup[subfigure]{font=scriptsize,labelformat=empty,justification=centering,singlelinecheck=false,skip=2pt}
  \begin{subfigure}[t]{0.238\textwidth}
    \centering
    \includegraphics[width=\linewidth]{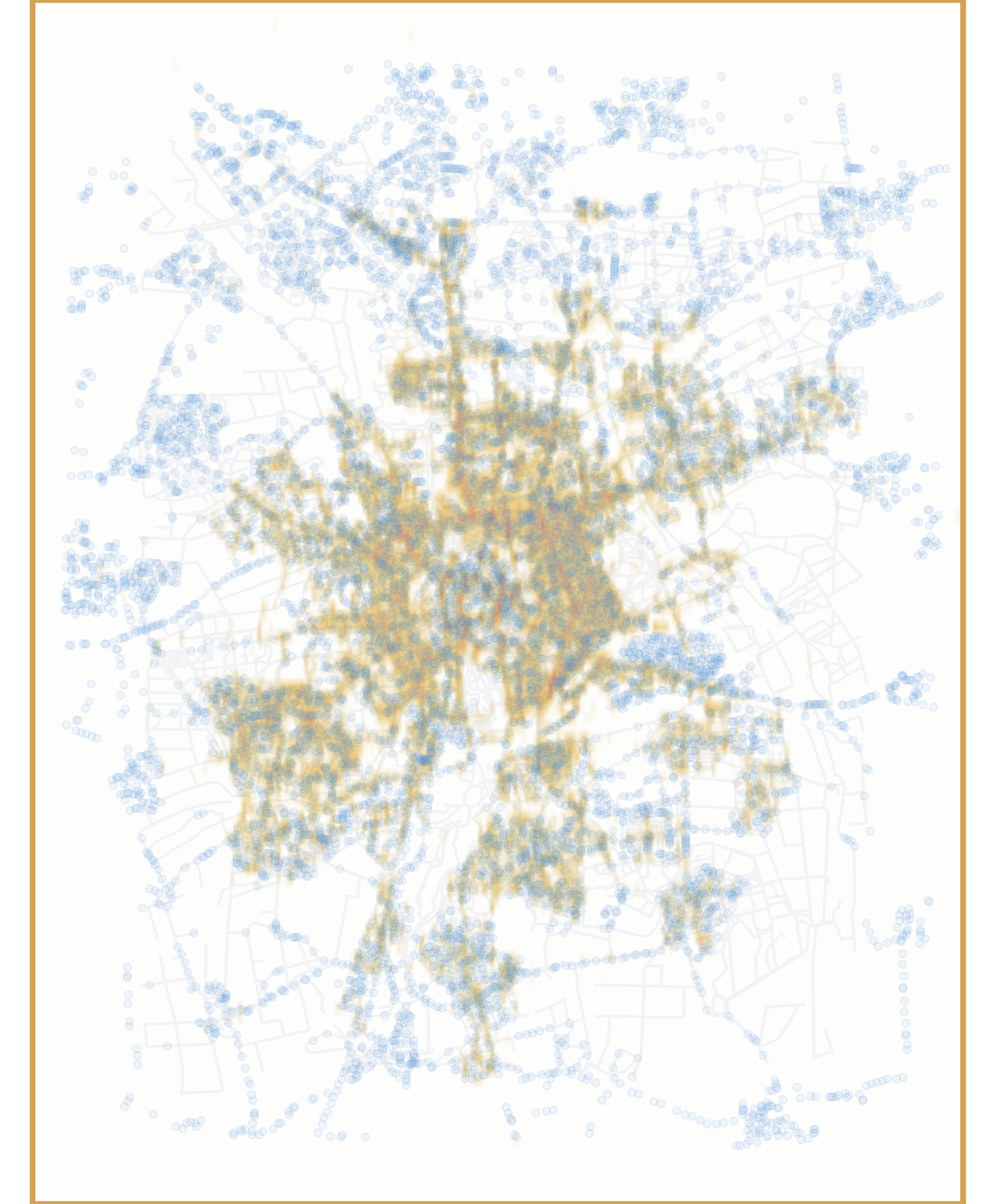}
    \caption{\textbf{A Demand/support}\\Trip endpoints and feasible support.}
  \end{subfigure}\hfill
  \begin{subfigure}[t]{0.238\textwidth}
    \centering
    \includegraphics[width=\linewidth]{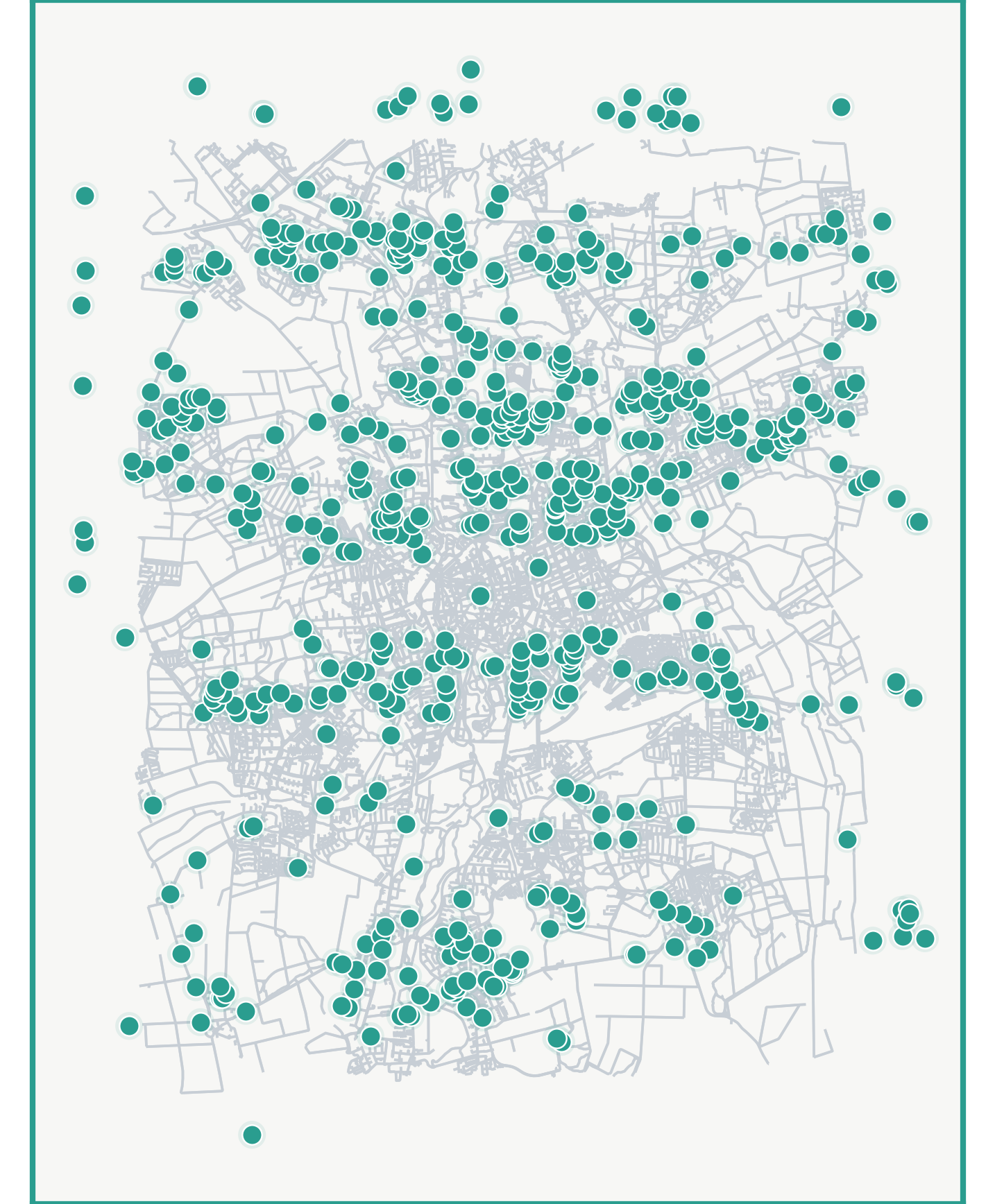}
    \caption{\textbf{B Baseline \(E_0\)}\\Selected parking-zone sites.}
  \end{subfigure}\hfill
  \begin{subfigure}[t]{0.238\textwidth}
    \centering
    \includegraphics[width=\linewidth]{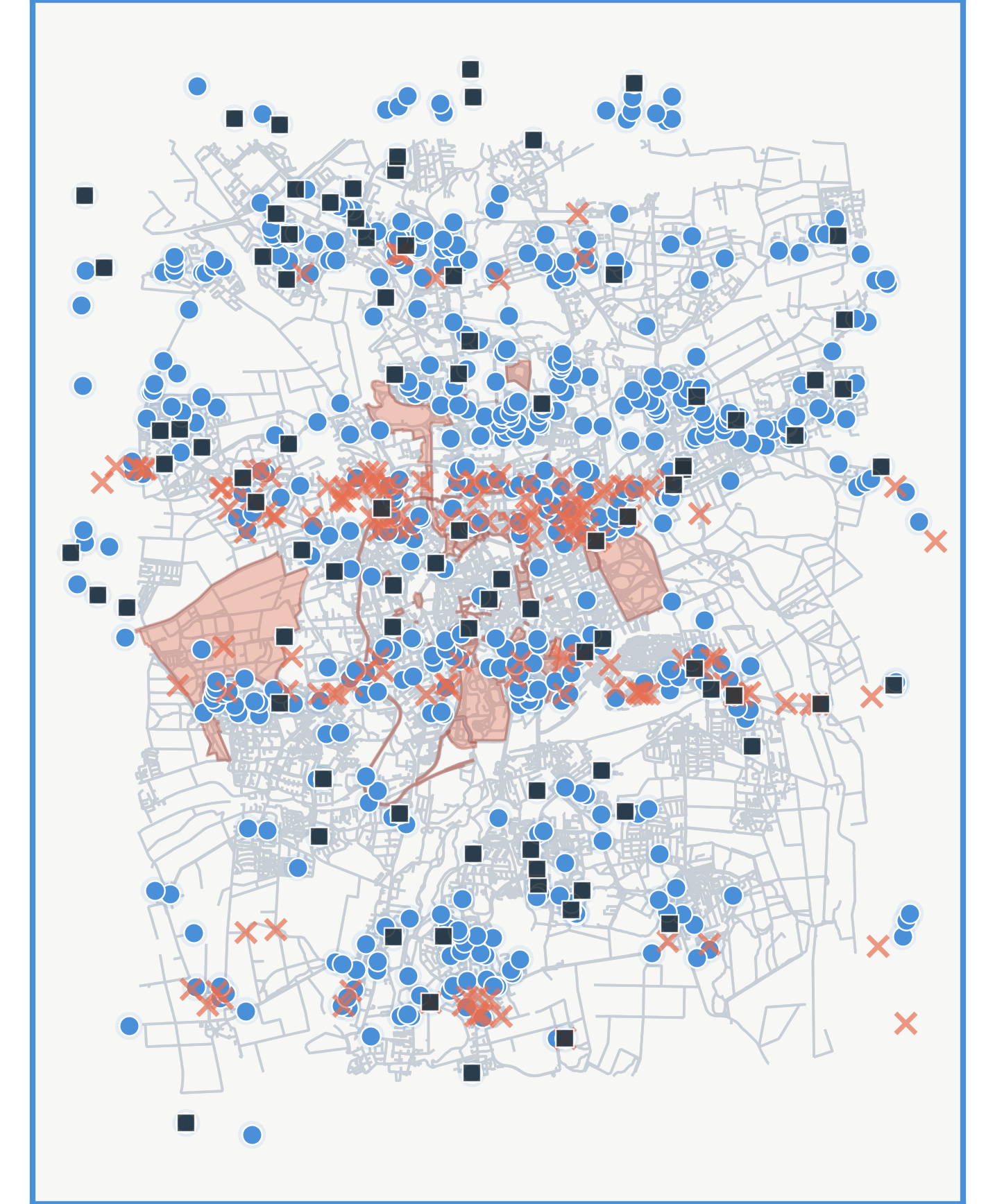}
    \caption{\textbf{C Edited \(E_5\)}\\Exclusions, locks, and spacing.}
  \end{subfigure}\hfill
  \begin{subfigure}[t]{0.238\textwidth}
    \centering
    \includegraphics[width=\linewidth]{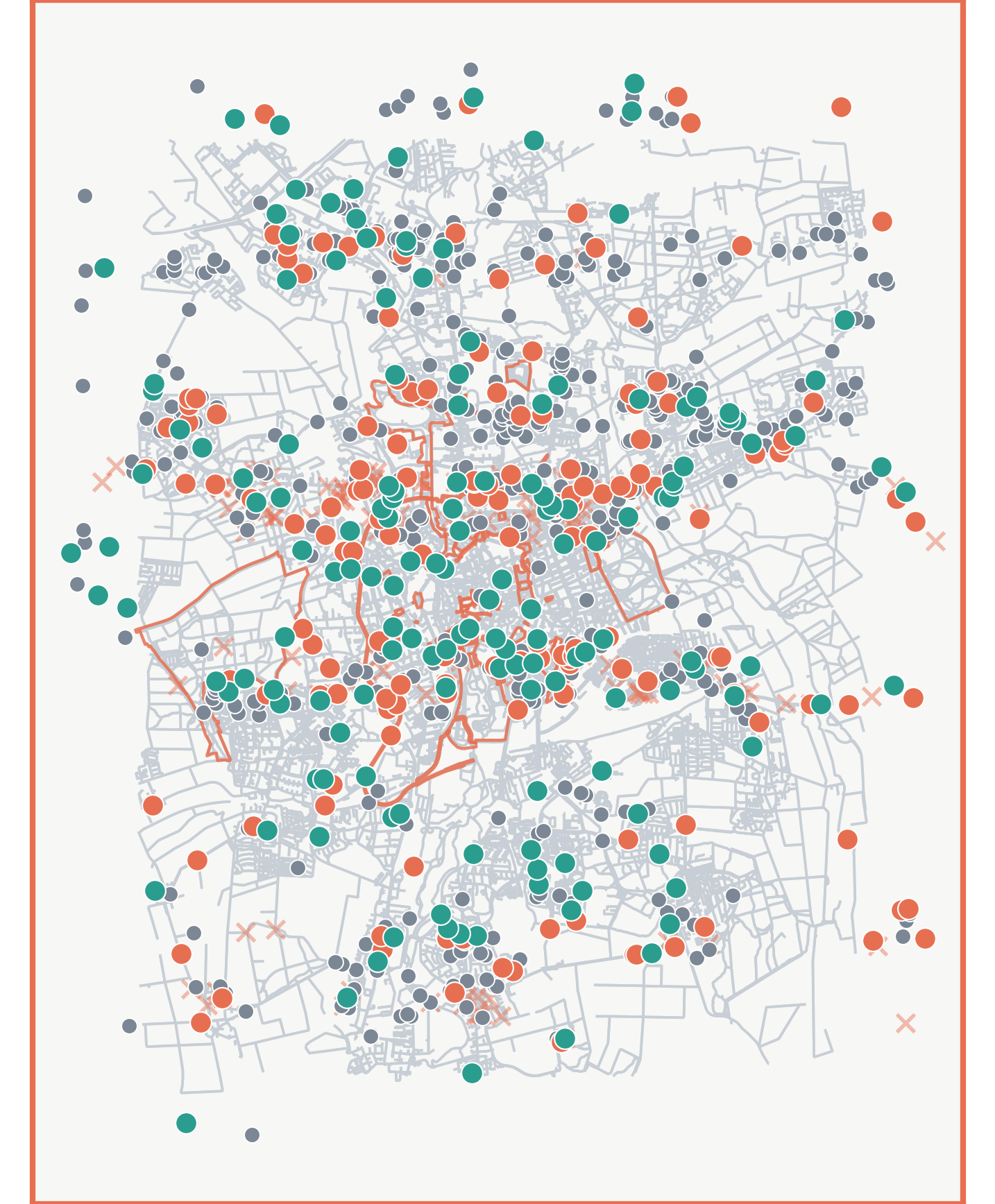}
    \caption{\textbf{D \(E_0\) to \(E_5\)}\\Retained, removed, and added sites.}
  \end{subfigure}
  \caption{Illustrative Braunschweig diagnostic on a separate \qty{25}{m} grid; quantitative
  comparisons use the \qty{10}{m} support in \cref{sec:evaluation-setup}. Panel A shows demand and
  candidates; B marks baseline sites in teal; C shows selected sites (blue), locks (dark squares), and
  exclusions (coral); D marks retained, removed, and added sites in gray, coral, and teal. At \(E_5\),
  \num{162} candidates are forbidden, \num{90} sites are locked, and \qty{25}{m} spacing is active;
  the plan retains \num{443} baseline sites and adds \num{157} replacements.}
  \Description{Four maps show demand and candidate support, the baseline selected plan, the edited
  plan with exclusions, locks, and spacing, and retained, removed, and added sites.}
  \label{fig:spatial-policy-edits}
\end{figure*}

CLIPPER solves each recorded policy state from the beginning; it carries neither selected sites nor
candidate scores over from the previous edit. Each primary comparison uses full-set greedy with the
same cap policy as the corresponding CLIPPER mode; we call this run the \emph{control}. Specifically,
CLIPPER-F and its control use balanced caps, while CLIPPER-A and its control use
coverage-prioritized caps. A coverage difference within either pair can therefore be attributed to
the restricted candidate pool rather than to different constraints. Every method is rerun on every
state in the complete \(E_0,\ldots,E_{10}\) chain (\cref{tab:method-comparison-main}).
\Cref{fig:spatial-policy-edits} shows one such edit on a separate illustrative Braunschweig
diagnostic: between \(E_0\) and \(E_5\), exclusions remove candidates, locks persist from the
baseline, and the replanned set retains most baseline sites while replacing others.

\subsection{Setup}
\label{sec:evaluation-setup}

\paragraph{Data construction.}
The benchmark retains every valid record in three non-public operator trip feeds available to the
project for internal shared-mobility research. For the paper-facing objective, origins and destinations
are snapped to spatial support and aggregated as endpoint counts; individual trip records,
coordinates, timestamps, and vehicle or provider identifiers remain non-public and are not released.
After validation, the feeds contain \num{243649} Braunschweig trips
(2024-01-01--2024-04-30), \num{4958963} Munich trips (2023-12-18--2024-08-22), and
\num{12306947} Berlin trips over the latter interval. We snap each valid origin and destination to
its nearest feasible support point and assign one unit of demand to each snapped endpoint; temporal
bins are summed for the paper's spatial objective.

Candidate grids use OpenStreetMap (OSM)-derived sidewalk, footway, and public-parking support
\cite{Haklay_Weber_2008} after restricted-area filtering, at \qty{10}{m} resolution in
Braunschweig and \qty{25}{m} in Munich and Berlin. We first fit \(k\)-means with 16 clusters (seed
\(42\)), then split or merge clusters until each proposal group contains
\(1{,}750\text{--}2{,}750\) candidate points. This yields 29 proposal groups over \num{60495}
unique candidates in Braunschweig, 16 groups over \num{32907} candidates in Munich, and 36 groups
over \num{68922} candidates in Berlin. \Cref{tab:benchmark-setup} summarizes the three instances.
The same synthetic partitions serve as proposal and accounting groups in
the benchmark. Each proposal group is therefore paired one-to-one with the accounting group that
contains its candidates, as required by CLIPPER-A's slot-allocation rule. The groups are synthetic
spatial partitions rather than administrative districts. No demand or candidate downsampling is
used.

\begin{table}[t]
  \centering
  \scriptsize
  \setlength{\tabcolsep}{2pt}
  \begin{tabular*}{\columnwidth}{@{\extracolsep{\fill}}lrrrrr@{}}
    \toprule
    City & Trips & Cand. & Budget & Groups & F slots \\
    \midrule
    Braunschweig & \num{243649} & \num{60495} & 600 & 29 & \num{29696} \\
    Munich & \num{4958963} & \num{32907} & 1200 & 16 & \num{16384} \\
    Berlin & \num{12306947} & \num{68922} & 1600 & 36 & \num{36864} \\
    \bottomrule
  \end{tabular*}
  \caption{Full-data instances. F slots are the maximum CLIPPER-F(\(K=1024\)) entries per round
  before filtering and deduplication; CLIPPER-A uses \(8192\) total slots.}
  \label{tab:benchmark-setup}
\end{table}

\paragraph{Spatial objective and constraints.}
All main-text methods use weighted point coverage within a \qty{200}{m} OSM walk-network radius.
For a given state, they share the complete demand and candidate data, encoded locks, exclusions,
conflict classes, spacing rules, and global facility budget. Coverage incidence and facility spacing
use shortest paths on per-city OSM walk graphs. In \cref{tab:method-comparison-main}, rows in the
balanced-cap block use equal caps for all accounting groups; rows in the coverage-prioritized block
use \(\lambda=1.5\) and leave the global budget unchanged. Each CLIPPER mode and its control use the
same caps.

\paragraph{Constructed edit chain.}
The \(E_0,\ldots,E_{10}\) chain is a controlled stress test whose complete configuration appears in
\cref{tab:scenario-chain}. Starting from the unrestricted \(E_0\) state, later states place increasing
shares of the most consequential baseline sites inside \qty{75}{m} exclusion zones and retain
increasing shares as mandatory locks. The locks form a growing prefix of baseline sites ordered by
their coverage contribution and spatial dispersion. From \(E_5\), the states add a minimum network
distance between selected sites. From \(E_8\), they also add \qty{150}{m} exclusion zones centered on
dense endpoint cells; the scenario records these halos as additional exclusions.

\input{tables/scenario_chain_table.tex}

\paragraph{Compared methods.}
The primary suite compares CLIPPER-F(\(1024\)) and
CLIPPER-A(\(B_{\mathrm{prop}}\in\{4096,8192\}\)) with full-set greedy under the same policy.
The auxiliary controls are competitive local search \cite{Lee_Sviridenko_Vondrak_2010}, stochastic
greedy \cite{Mirzasoleiman_2015}, and thresholded greedy \cite{Badanidiyuru_Vondrak_2014}.

\paragraph{Audit of the fixed candidate order.}
For CLIPPER-F, a separate exact audit scans every remaining feasible candidate after each selection
for \(K\in\{256,512,1024\}\), in all 11 states and without downsampling. It also performs one
terminal check when the pool has no positive feasible action. Let \(\mathcal T_{\mathrm{audit}}\)
contain these audited rounds. In round \(t\), \(m_t^*\) is the largest feasible gain found by the
full scan, while \(m_t\) is the largest gain among candidates in the pool (\cref{sec:method}). We
record whether their difference is at most
\(\max(10^{-9},10^{-6}|m_t^*|)\). The reported percentage is therefore the share of audited rounds in
which the pool omits no larger feasible gain within this tolerance, including terminal checks with
both gains zero. This compares gains, not candidate identities. We also sum \(\delta_t=m_t^*-m_t\) over
\(\mathcal T_{\mathrm{audit}}\) and divide by the coverage added after mandatory locks. This second
quantity reports the size of the gains omitted in individual rounds relative to the coverage added
by the run.

\paragraph{Comparison with scores updated after each selection.}
The fixed rule computes singleton coverage once at the start of a state. The comparison rule starts
from the same scores, builds an index from each demand point to the candidates that cover it, and
updates only candidates affected by demand that has just become covered. The heap for each group then
exposes its current top \(K=1024\) candidates. Both rules use the same Python executor, which selects
one candidate per step, as well as the same data, policy checks, and stopping rule. Their timers
include score initialization, updates, heap
maintenance, pool construction, and selection; audits are disabled. These paired runs use the same
host as the primary comparison. Panel~(b) of
\cref{tab:proposal-order-audit} compares the two rules within this executor, whereas panel~(a)
reports the separate compiled exact-audit execution. Their fixed-order coverage values differ
slightly; the paired comparison uses only panel~(b)'s fixed-order reference.

\paragraph{Documented municipal policy inputs.}
We additionally convert two official Braunschweig open-data resource distributions of
e-scooter no-parking zones, dated 2025-01-28 and 2026-02-17, into exclusion-only states
\cite{StadtBraunschweig_Parkverbotszonen_2026}. The CC BY 4.0 archives are pinned by resource ID and
SHA-256 digest. We repair and union the polygons in EPSG:25832, reproject the complete
\num{60495}-candidate support, and mark every intersecting candidate as forbidden. These standalone
snapshots use the same demand and budget with no locks or spacing, isolating how the documented
policy geometries change feasibility and the returned plan.

\paragraph{Metrics and timing boundary.}
Let \(W=\sum_{d\in\mathcal D}w_d\) be total demand weight. Coverage percentage is
\(100f(S)/W\): the share of trip endpoints that lie within the walk radius of some selected site.
Coverage gaps subtract method coverage from the corresponding control. A gap of
\(0.245\) percentage points means that the plan covers \(0.245\) percentage points less of total
demand weight than the control. Thus the gap measures agreement with an empirical greedy reference
rather than global optimality. All primary and paired score-update timings use the same
Intel Core Ultra 9 285K host. In \cref{tab:method-comparison-main}, CLIPPER and full-set greedy
share the rollout boundary: the timer starts after scenario initialization and
includes per-round pool construction, exact marginal evaluation, feasibility checks, and selection.
It excludes scenario construction, preliminary validation of candidate-pool dimensions, raw
preprocessing, optional screening and exact auditing, and output serialization. Optimization and
mixed-integer linear programming (MILP) diagnostics report their own broader boundaries separately.

\subsection{Results}

\subsubsection{Full-city comparison under the same policy (Q1)}
\Cref{tab:method-comparison-main} reports means across all 11 states. The mean coverage gap for
CLIPPER-F(\(1024\)) is \(0.245\) percentage points in Braunschweig, \(0.003\) in Munich, and
\(0.001\) in Berlin. Mean rollout time falls from \(24.029\) to \(1.762\) seconds in
Braunschweig, \(22.943\) to \(1.493\) in Munich, and \(52.685\) to \(1.825\) in Berlin. The
corresponding speedups are \(13.63\times\) in Braunschweig, \(15.37\times\) in Munich, and
\(28.87\times\) in Berlin.

\input{tables/point_method_comparison_main_table.tex}

\input{tables/point_proposal_order_audit_table.tex}

Summed over the 11 states of one city, the tabulated means correspond to
\(16.4\text{--}20.1\) seconds for CLIPPER-F and \(4.2\text{--}9.7\) minutes for its control
(\cref{tab:method-comparison-main}). Per-state results show that these speedups persist across the
chain. The coverage gaps range from \(-0.038\) to \(0.377\) percentage points in Braunschweig,
from \(-0.010\) to \(0.016\) in Munich, and from \(-0.001\) to \(0.002\) in Berlin. The corresponding
speedups range from \(7.4\) to \(40.7\times\), from \(8.4\) to \(49.1\times\), and from \(15.7\) to
\(84.3\times\), respectively. Speedups are largest before spacing activates at \(E_5\), but remain
at least \(7.4\times\) once spacing and the exclusion halos are
combined.
The small negative gaps occur because the two greedy runs can follow different selection sequences
and end in different feasible plans.

\paragraph{CLIPPER-A}
Under the coverage-prioritized cap policy, CLIPPER-A(\(8192\)) has gaps to its control of \(1.82\)
percentage points in Braunschweig, \(0.12\) in Munich, and \(0.27\) in Berlin. Its rollout time is
\(0.11\), \(0.15\), and \(0.09\) times the corresponding control, respectively. CLIPPER-A and
CLIPPER-F differ in both cap policy and pool allocation, so differences between the modes reflect
both choices.

\paragraph{Controls and optimization checks.}
Two additional cap-policy ablations support this reading: adaptive pool budgets under balanced caps
remain near the control, and under coverage-prioritized caps fixed per-group pools underperform
CLIPPER-A(\(8192\)) by roughly \(14.1\) percentage points in Braunschweig, \(2.8\) in Munich, and
\(4.9\) in Berlin. Competitive local search reaches similar
coverage but requires \(22.4\text{--}72.3\) seconds under its
reported boundary; stochastic and thresholded variants are slower, lower-quality, or both. Further
supporting diagnostics are exact \(E_0\) solves with small facility budgets and mixed-integer solves
over the restricted pool with the full facility budget. They distinguish closeness to full-set
greedy from closeness to an optimum.

\subsubsection{A fixed candidate order and scores updated during selection (Q2)}
CLIPPER-F ranks each group's candidates once per state by singleton coverage and then reuses that
order. As sites are added, candidates cover overlapping demand and their current gains change. The
audit in \cref{tab:proposal-order-audit} asks whether the restricted pool omits any larger feasible
gain found by scanning all remaining candidates. With \(K=256\), no such gain is omitted
in \(4.4\%\) of audited rounds in Braunschweig, \(10.5\%\) in Munich, and \(24.4\%\) in
Berlin. The percentages rise to \(13.3\%\), \(66.4\%\), and \(95.6\%\) at \(K=512\), and to
\(51.0\%\), \(100.0\%\), and \(100.0\%\) at \(K=1024\), respectively. At \(K=1024\), no audited run
terminates while a candidate outside the pool could still increase coverage. The sum of omitted gains
divided by coverage added after locks is \(0.140\) in Braunschweig and \(0\) in Munich and Berlin.

The round-by-round percentages and final coverage answer different questions. In Braunschweig, the
\(K=1024\) pool omits no larger feasible gain in \(51\%\) of audited rounds,
yet its mean final coverage over the eleven states is only \(0.245\) percentage points below the
full-set control. A later
selection can cover demand omitted in an earlier round. The result shows that an order computed once
can cease to identify the largest current gains without causing an early stop or a similarly large
change in final coverage.

Within the stepwise executor used for the paired comparison, updating affected scores changes mean
coverage by \(+0.2037\) percentage points in Braunschweig, \(+0.0049\) in Munich, and \(-0.0006\)
in Berlin.
Braunschweig is the only city that changes by more than \(0.01\) percentage points and the only one
where the exact audit finds omitted gains at \(K=1024\). After initialization, the update rule touches
only \(0.128\%\) of stored candidate rows in Braunschweig, \(0.050\%\) in Munich, and \(0.026\%\) in
Berlin. Maintaining and reading the heaps raises the median paired wall-time ratio to \(2.75\times\),
\(1.76\times\), and \(2.32\times\), respectively. Updating scores thus
provides a measurable coverage benefit in Braunschweig and little or none in the other two cities,
while requiring more time in all three (\cref{tab:proposal-order-audit}).

\subsubsection{Documented policy-input case (Q3)}
\Cref{tab:policy-snapshots} uses two official resource versions instead of constructed exclusions.
The polygon union grows by \qty{3.154}{\square\kilo\metre}; the number of candidates inside an
exclusion grows from \num{1435} (\(2.37\%\)) to \num{4653} (\(7.69\%\)). This yields
\num{3218} newly forbidden candidates and no candidate that becomes feasible again.

Both rules stop below the 600-site budget because the terminal full-set check finds no positive
feasible gain: after 540 selections with fixed scores and 541 with updated scores. With fixed scores,
CLIPPER-F retains 510 of 540 selected sites; with updated scores, it retains 511 of 541. Both rules
replace 30 sites: 22 were directly
made infeasible by the newer resource geometry, while eight remained
feasible and changed later in the selection sequence. Coverage changes by \(-0.094\) and \(-0.106\)
percentage points, respectively. Similar coverage therefore does not mean a similar plan: judged by
the objective alone the two official policy versions look almost interchangeable, yet their plans
differ in 30 of about \num{540} sites that still require municipal assessment, implementation, and
communication.

\input{tables/brunswick_policy_snapshot_table.tex}

\subsubsection{Recorded audit and repeatability checks}
\label{sec:cert-empirical}
For each audited round, the implementation computes the largest feasible gain in the full remaining
set, the largest gain in the restricted pool, their difference, and the terminal check. Panel~(b)
disables these scans so that it measures only the two candidate-ordering rules in the stepwise
executor.

In a separate full-data diagnostic of CLIPPER-F(\(1024\)) using the non-public inputs, three reruns
of every state and city (\(99\) runs) produced identical coverage, step counts, numbers of gain
calculations, termination reasons, and terminal flags within each state. Every run ended with no
feasible candidate that could add coverage, and none stopped because its candidate pool was too
narrow. These checks and the municipal source manifest are retained with the project records.

%% file: tables/scenario_chain_table.tex
\begin{table*}[t]
  \centering
  \scriptsize
  \setlength{\tabcolsep}{3.6pt}
  \renewcommand{\arraystretch}{1.02}
  \begin{tabular*}{\textwidth}{@{\extracolsep{\fill}}lrrrrrrrrrrr@{}}
    \toprule
    State & \(E_0\) & \(E_1\) & \(E_2\) & \(E_3\) & \(E_4\) & \(E_5\) & \(E_6\) & \(E_7\) & \(E_8\) & \(E_9\) & \(E_{10}\) \\
    \midrule
    Baseline sites inside \qty{75}{m} exclusions (\%) & 0.0 & 2.5 & 5.0 & 5.0 & 7.5 & 7.5 & 10.0 & 12.5 & 12.5 & 15.0 & 17.5 \\
    Baseline sites kept as locks (\%)                  & 0.0 & 0.0 & 0.0 & 10.0 & 10.0 & 15.0 & 20.0 & 20.0 & 25.0 & 30.0 & 35.0 \\
    \qty{150}{m} exclusion halos                       & 0 & 0 & 0 & 0 & 0 & 0 & 0 & 0 & 10 & 15 & 20 \\
    Minimum network spacing (m)                        & 0 & 0 & 0 & 0 & 0 & 25 & 35 & 45 & 50 & 55 & 60 \\
    \bottomrule
  \end{tabular*}
  \caption{Complete configuration of the controlled edit chain. Exclusions target a growing prefix
  of baseline sites ordered by individual coverage contribution; locks use contribution and spatial
  dispersion. Each halo is an exclusion zone centered on a dense endpoint cell.}
  \label{tab:scenario-chain}
\end{table*}

%% file: tables/point_method_comparison_main_table.tex
\begin{table*}[t]
  \centering
  \scriptsize
  \setlength{\tabcolsep}{3.0pt}
  \renewcommand{\arraystretch}{1.08}
  \begin{tabular*}{\textwidth}{@{\extracolsep{\fill}}lrrrrrrrrrrrr@{}}
    \toprule
    Method / control
      & \multicolumn{3}{c}{Mean coverage (\%)}
      & \multicolumn{3}{c}{Mean gap to control (pp)}
      & \multicolumn{3}{c}{Mean rollout (s)}
      & \multicolumn{3}{c}{Rollout/control} \\
    \cmidrule(lr){2-4}\cmidrule(lr){5-7}\cmidrule(lr){8-10}\cmidrule(l){11-13}
      & BS & MUC & BER & BS & MUC & BER & BS & MUC & BER & BS & MUC & BER \\
    \midrule
    \multicolumn{13}{@{}l}{\textit{Balanced caps: full-set greedy and CLIPPER-F}} \\
    Full-set greedy & 90.4 & 69.7 & 59.5 & 0.000 & 0.000 & 0.000 & 24.029 & 22.943 & 52.685 & 1.000 & 1.000 & 1.000 \\
    CLIPPER-F(1024) & 90.1 & 69.7 & 59.5 & 0.245 & 0.003 & 0.001 & 1.762 & 1.493 & 1.825 & 0.073 & 0.065 & 0.035 \\
    \midrule
    \multicolumn{13}{@{}l}{\textit{Coverage-prioritized caps: full-set greedy and CLIPPER-A}} \\
    Full-set greedy & 96.8 & 81.2 & 69.5 & 0.000 & 0.000 & 0.000 & 28.343 & 22.931 & 48.826 & 1.000 & 1.000 & 1.000 \\
    CLIPPER-A(4096) & 89.8 & 78.2 & 66.1 & 6.920 & 3.010 & 3.410 & 3.222 & 3.383 & 4.393 & 0.114 & 0.148 & 0.090 \\
    CLIPPER-A(8192) & 95.0 & 81.1 & 69.3 & 1.820 & 0.120 & 0.270 & 3.219 & 3.360 & 4.415 & 0.114 & 0.147 & 0.090 \\
    \bottomrule
  \end{tabular*}
  \caption{Main comparison on complete data, averaged over \(E_0,\ldots,E_{10}\). BS, MUC, and BER
  denote the three cities; pp denotes percentage points. Each block pairs CLIPPER with full-set greedy
  under the same caps, on the same host and with the same rollout timing boundary
  (\cref{sec:evaluation-setup}). Gap is control minus method coverage; rollout/control is their time ratio.}
  \label{tab:method-comparison-main}
\end{table*}

%% file: tables/point_proposal_order_audit_table.tex
\begin{table*}[t]
  \centering
  \scriptsize
  \setlength{\tabcolsep}{4pt}
  \begin{tabular*}{\textwidth}{@{\extracolsep{\fill}}llrrrr@{}}
    \toprule
    \multicolumn{6}{l}{\textbf{(a) Audit of a candidate order computed once per state}} \\
    City & \(K\) & \shortstack{Mean\\coverage (\%)} &
    \shortstack{Audited rounds: no larger\\feasible gain omitted (\%)} &
    \shortstack{Omitted gains /\\added coverage} &
    \shortstack{Runs stopped by\\pool limit (of 11)} \\
    \midrule
    Braunschweig & 256  & 77.248 & 4.4   & 3.295 & 11/11 \\
                 & 512  & 85.677 & 13.3  & 1.071 & 11/11 \\
                 & 1024 & 90.132 & 51.0  & 0.140 & 0/11  \\
    Munich       & 256  & 64.623 & 10.5  & 0.748 & 6/11  \\
                 & 512  & 69.395 & 66.4  & 0.046 & 0/11  \\
                 & 1024 & 69.685 & 100.0 & 0.000 & 0/11  \\
    Berlin       & 256  & 58.638 & 24.4  & 0.214 & 0/11  \\
                 & 512  & 59.542 & 95.6  & 0.002 & 0/11  \\
                 & 1024 & 59.546 & 100.0 & 0.000 & 0/11  \\
    \bottomrule
  \end{tabular*}

  \vspace{3pt}
  \begin{tabular*}{\textwidth}{@{\extracolsep{\fill}}lrrrrrrr@{}}
    \toprule
    \multicolumn{8}{l}{\textbf{(b) Scores fixed at state start versus scores updated after each selection, \(K=1024\)}} \\
    City & \shortstack{Fixed scores:\\mean cov. (\%)} &
    \shortstack{Updated scores:\\mean cov. (\%)} &
    \shortstack{Coverage\\change (pp)} &
    \shortstack{Fixed scores:\\median wall (s)} &
    \shortstack{Updated scores:\\median wall (s)} &
    \shortstack{Updated / fixed\\wall time} &
    \shortstack{Score rows\\updated (\%)} \\
    \midrule
    Braunschweig & 90.172 & 90.375 & \(+0.2037\) & 24.04 & 61.08  & 2.75 & 0.128 \\
    Munich       & 69.678 & 69.683 & \(+0.0049\) & 39.09 & 74.73  & 1.76 & 0.050 \\
    Berlin       & 59.547 & 59.546 & \(-0.0006\) & 68.75 & 152.85 & 2.32 & 0.026 \\
    \bottomrule
  \end{tabular*}
  \caption{Results on \(E_0,\ldots,E_{10}\). Panel~(a) scans the full remaining set after each
  selection and at termination; it reports rounds with no larger feasible gain omitted, normalized omitted
  gains, and stops caused by the pool limit. Panel~(b) uses the stepwise executor from
  \cref{sec:evaluation-setup}; its timer includes initialization, updates, heap maintenance, pool
  construction, and selection. The host is shared with \cref{tab:method-comparison-main}, but the
  execution path and timer differ; panel~(b)'s times are compared only within that panel.
  Coverage values are means; wall times and paired per-state time ratios are medians.
  The final column normalizes post-initialization score updates by the product of
  selection calls and stored candidate rows.}
  \label{tab:proposal-order-audit}
\end{table*}

%% file: tables/brunswick_policy_snapshot_table.tex
\begin{table}[!b]
  \centering
  \scriptsize
  \setlength{\tabcolsep}{2.5pt}
  \begin{tabular*}{\columnwidth}{@{\extracolsep{\fill}}lrrrrr@{}}
    \toprule
    \multicolumn{6}{l}{\textbf{(a) Policy resources and resulting coverage}} \\
    Resource date & Zones & Area (\(\mathrm{km}^2\)) & \shortstack[r]{Forbidden\\cand.} &
      \shortstack[r]{Fixed scores:\\cov. (\%)} & \shortstack[r]{Updated scores:\\cov. (\%)} \\
    \midrule
    2025-01-28 & 51 & 6.275 & \num{1435} & 91.609 & 91.807 \\
    2026-02-17 & 55 & 9.429 & \num{4653} & 91.515 & 91.701 \\
    \bottomrule
  \end{tabular*}

  \vspace{3pt}
  \begin{tabular*}{\columnwidth}{@{\extracolsep{\fill}}lrrrrr@{}}
    \toprule
    \multicolumn{6}{l}{\textbf{(b) Change in selected sites from the first resource to the second}} \\
    Candidate scores & \shortstack{Selected\\old/new} & Retained & Removed &
      \shortstack{Removed after becoming\\newly forbidden} & Added \\
    \midrule
    Fixed at state start & 540/540 & 510 & 30 & 22 & 30 \\
    Updated after selection & 541/541 & 511 & 30 & 22 & 30 \\
    \bottomrule
  \end{tabular*}
  \caption{Braunschweig case with two official no-parking-zone resources and all \num{60495}
  candidates. Both runs use \(K=1024\), identical demand, a 600-site budget, and no locks or spacing;
  of 30 removed sites, 22 became forbidden and eight remained feasible.}
  \label{tab:policy-snapshots}
\end{table}

%% file: sections/implementation_pathway.tex
\section{Municipal implementation pathway}
\label{sec:implementation}

We propose a municipal process for comparing policy alternatives using CLIPPER's optimization
and audit functions (\cref{tab:implementation-pathway}). In the Braunschweig case, the plans differ
in 30 sites for only about \(0.1\) percentage points of coverage, so the interface must show site
changes beside coverage. At \(K=1024\), the pool omits no larger feasible gain in every audited
Munich and Berlin round but in \(51\%\) of Braunschweig rounds; planners therefore need to
calibrate score updates locally (\cref{tab:proposal-order-audit,tab:policy-snapshots}). To support
this calibration and the final decision, we propose scanning every remaining candidate and
recording what the pool omitted.

\input{tables/municipal_implementation_pathway_table.tex}

The collaboration established the workflow requirements. In the Braunschweig case, we registered
two official policy versions and compared the resulting plans (\cref{tab:policy-snapshots}). In the proposed
pilot, a municipal policy owner
defines constraints and acceptance checks, and a data steward
validates demand, candidates, and the network. Each edit produces a fresh scenario
\(\Omega\); the interface shows coverage and the sites retained, removed, or added. The chosen
state then receives the final scan and archive package in
\cref{tab:implementation-pathway}. Following planning-support practice
\cite{Pelzer_2014,Russo_2018}, evaluation should compare full-set and CLIPPER-assisted sessions and
measure time to an accepted scenario, alternatives inspected, constraint corrections, final
coverage, and the usefulness of the change and audit records.

%% file: tables/municipal_implementation_pathway_table.tex
\begin{table}[!b]
  \centering
  \scriptsize
  \setlength{\tabcolsep}{3pt}
  \renewcommand{\arraystretch}{1.05}
  \begin{tabular}{@{}p{0.18\columnwidth}p{0.78\columnwidth}@{}}
    \toprule
    Stage & Inputs, outputs, and acceptance check \\
    \midrule
    Register & \textit{Input:} policy geometry, candidates, demand, coordinate system, validity date.
      \textit{Check:} source checksum, valid geometry, and candidate intersections. \\
    Calibrate & \textit{Input:} representative recorded states and local time/quality tolerances.
      \textit{Check:} choose \(K\), cap policy, and candidate rule; meet local targets and scan for
      improving candidates outside the pool when a run stops. \\
    Compare & \textit{Input:} exclusions, locks, spacing, caps, and facility budget for each
      alternative. \textit{Check:} feasible plan, coverage, changed sites, terminal status, and
      input/output checksums. \\
    Final decision & \textit{Input:} chosen policy state and audit level. \textit{Output and check:}
      scan all remaining candidates, archive inputs and outputs, and obtain policy-owner sign-off. \\
    \bottomrule
  \end{tabular}
  \caption{Municipal pilot pathway. The municipality owns policy and acceptance thresholds; CLIPPER
  supplies plans, comparisons, and audit records.}
  \label{tab:implementation-pathway}
\end{table}

%% file: sections/discussion.tex
\section{Limitations}
\label{sec:discussion}

\paragraph{Evidence about planning use.}
The Braunschweig collaboration informed requirements; the experiments evaluate optimizer behavior
rather than user interaction or deployment outcomes.
Without recorded planning sessions or edit logs, the study does not estimate edit frequency,
acceptable response times, alternatives considered, or planner utility. Reported speedups cover
optimization rather than end-to-end deliberation.

\paragraph{Model, data, and transfer.}
The experiments use the same partitions for proposal groups and accounting caps; overlapping
proposal groups were not evaluated.
The weighted point-coverage model does not represent curb-segment capacity, facility congestion,
user compliance, equity outcomes, temporal rebalancing, operator behavior, or institutional
approval. Addressing these dimensions requires assignment, occupancy, or additional utility and
feasibility inputs. The three trip feeds are non-public, incomplete market samples, and a held-out
city--provider pair is needed to assess transfer beyond them. Independent reconstruction additionally
requires the processed candidates, recorded scenarios, and fixed OSM snapshots; aggregate paper
tables alone are insufficient. In this paper, replay means deterministic execution of a versioned
scenario bundle within its recorded software environment. It is distinct from public reproduction
because the operator trip feeds cannot be released.

\paragraph{Runtime comparisons.}
Speedups compare implementations within the stated timing boundary, without separating pool
restriction from execution technique. A comparison with classical
lazy greedy under matched execution and timing conditions remains open.

\paragraph{Candidate scores.}
Both the fixed and updated candidate orders restart after each edit (\cref{tab:proposal-order-audit}). Learned or
capacity-aware candidate rules can use the same selector, but require validation on unseen data and
checks for bias \cite{Mehrabi_2022}.

%% file: sections/conclusion.tex
\section{Conclusion}

Across the three cities, CLIPPER-F's mean coverage gap over the complete chain is at most \(0.245\)
percentage points while its mean rollout is \(13.6\text{--}28.9\times\) faster than full-set greedy
under the same policy (\cref{tab:method-comparison-main}). Both CLIPPER modes enforce every encoded
hard constraint and replay recorded states deterministically. At \(K=1024\), none of the 11 audited
runs per city ends because of the pool limit, while the score-update comparison shows the local
trade-off between coverage and computation time (\cref{tab:proposal-order-audit}).

These functions serve distinct planning tasks: the optimizer computes feasible alternatives, the
audit measures omitted gains, and comparisons between policy versions identify changed sites.
The Braunschweig plans differ in 30 sites despite similar coverage (\cref{tab:policy-snapshots}),
so aggregate coverage alone cannot describe what a municipality would implement. We propose using
these functions to compare alternatives, calibrate score updates locally, and scan the full candidate
set before a final decision. The archive retains the policy inputs and selected sites.

\begin{acks}
We thank our partners at the City of Braunschweig for contributing operational requirements.
Generative AI tools were used only for editing and language polishing. The authors take full
responsibility for the scientific content, experiments, analyses, and conclusions.
\end{acks}